\documentclass[conference]{IEEEtran}
\IEEEoverridecommandlockouts
\usepackage{cite}
\usepackage{amsmath,amssymb,amsfonts}
\usepackage{algorithmic}
\usepackage{graphicx}
\usepackage{textcomp}
\usepackage{xcolor}
\usepackage{booktabs}
\usepackage{makecell}
\usepackage{colortbl}
\usepackage{color}
\definecolor{mygray}{gray}{.9}

\def\BibTeX{{\rm B\kern-.05em{\sc i\kern-.025em b}\kern-.08em
    T\kern-.1667em\lower.7ex\hbox{E}\kern-.125emX}}
\begin{document}

\title{ViDTA: Enhanced Drug-Target Affinity Prediction via Virtual Graph Nodes and Attention-based Feature Fusion\\
\thanks{Minghui’s work is supported in part by the National Natural Science Foundation of China (Grant No. 62202186). Shengshan’s work is supported in part by the National Natural Science Foundation of China (Grant No.62372196). Shengqing’s work is supported in part by the Hubei Provincial Natural Science Foundation Project(Grant No. 2023AFB342)and the Open Program of Nuclear Medicine and Molecular Imaging Key Laboratory of Hubei Province (Grant No. 2022fzyx018). The work is supported by the HPC Platform of Huazhong University of Science and Technology.\\
\indent  Shengqing is the corresponding author.}
}

\author{
\IEEEauthorblockN{1\textsuperscript{st} Minghui Li}
\IEEEauthorblockA{\textit{School of Software Engineering} \\
\textit{Huazhong University of}\\
\textit{Science and Technology}\\
Wuhan, China \\
minghuili@hust.edu.cn}
\and
\IEEEauthorblockN{2\textsuperscript{nd} Zikang Guo}
\IEEEauthorblockA{\textit{School of Software Engineering} \\
\textit{Huazhong University of}\\
\textit{Science and Technology}\\
Wuhan, China \\
zikangguo@hust.edu.cn}
\and
\IEEEauthorblockN{3\textsuperscript{rd} Yang Wu}
\IEEEauthorblockA{\textit{School of Software Engineering} \\
\textit{Huazhong University of}\\
\textit{Science and Technology}\\
Wuhan, China \\
yungwu@hust.edu.cn}
\and
\IEEEauthorblockN{4\textsuperscript{th} Peijin Guo}
\IEEEauthorblockA{
\textit{School of Cyber Science} \\
\textit{and Engineering} \\
\textit{Huazhong University of}\\
\textit{Science and Technology}\\
Wuhan, China \\
gpj@hust.edu.cn}
\and
\IEEEauthorblockN{5\textsuperscript{th} Yao Shi}
\IEEEauthorblockA{\textit{School of Software Engineering} \\
\textit{Huazhong University of}\\
\textit{Science and Technology}\\
Wuhan, China \\
yaoshi@hust.edu.cn}
\and
\IEEEauthorblockN{6\textsuperscript{th} Shengshan Hu}
\IEEEauthorblockA{
\textit{School of Cyber Science} \\
\textit{and Engineering} \\
\textit{Huazhong University of}\\
\textit{Science and Technology}\\
Wuhan, China \\
hushengshan@hust.edu.cn}
\and
\IEEEauthorblockN{7\textsuperscript{th} Wei Wan}
\IEEEauthorblockA{
\textit{School of Cyber Science} \\
\textit{and Engineering} \\
\textit{Huazhong University of}\\
\textit{Science and Technology}\\
Wuhan, China \\
wanwei\_0303@hust.edu.cn}
\and
\IEEEauthorblockN{8\textsuperscript{th} Shengqing Hu}
\IEEEauthorblockA{
\textit{Union Hospital,} \\
\textit{Tongji Medical College} \\
\textit{Huazhong University of}\\
\textit{Science and Technology}\\
Wuhan, China \\
hsqha@126.com}
}

\maketitle

\begin{abstract}
Drug-target interaction is fundamental in understanding how drugs affect biological systems, and accurately predicting drug-target affinity (DTA) is vital for drug discovery. Recently, deep learning methods have emerged as a significant approach for estimating the binding strength between drugs and target proteins. However, existing methods simply utilize the drug's local information from molecular topology rather than global information. Additionally, the features of drugs and proteins are usually fused with a simple concatenation operation,  limiting their effectiveness. To address these challenges, we proposed ViDTA, an enhanced DTA prediction framework. We introduce virtual nodes into the Graph Neural Network (GNN)-based drug feature extraction network, which acts as a global memory to exchange messages more efficiently. By incorporating virtual graph nodes, we seamlessly integrate local and global features of drug molecular structures, expanding the GNN's receptive field. Additionally, we propose an attention-based linear feature fusion network for better capturing the interaction information between drugs and proteins. Experimental results evaluated on various benchmarks including Davis, Metz, and KIBA  demonstrate that our proposed ViDTA outperforms the state-of-the-art baselines.

\end{abstract}


\section{Introduction}
The binding affinity between drugs and the target proteins plays an important role in numerous biological processes, such as immune responses \cite{b1} and gene regulation \cite{b2}. Traditional high-throughput screening experiments for measuring affinity are labor-intensive, time-consuming, and expensive \cite{b3}. Therefore, computational methods for predicting drug-target affinity (DTA) have emerged.

Traditional machine learning methods, such as Support Vector Machines (SVM) \cite{b4} and Random Forests (RF) \cite{b5}, have been widely applied to DTA prediction. However, these methods require complex and time-consuming feature engineering and suffer from low prediction accuracy due to limited and non-uniform datasets.
Recently, deep learning has emerged as a promising solution to deal with protein structures \cite{b6, b7, b8}. 
Unfortunately, it is unsuitable for DTA prediction due to the lack of structural information in drug-target samples. 
Although we can exploit deep learning methods to predict structures, they often introduce accumulated errors, reducing prediction accuracy.

In light of this, recent works turn to developing end-to-end models that only take the drug’s SMILES sequence and the protein’s amino acid sequence as input. For instance, DeepDTA \cite{b9} utilizes 1D-CNNs to extract features from both drug and protein sequences to predict DTA. AttentionDTA \cite{b10} employs a bidirectional multi-head attention mechanism to highlight key subsequences in drug and protein sequences. TEFDTA \cite{b11} uses Transformers and 1D-CNNs to extract drug and protein features. However, these methods failed to examine the critical role of atomic properties and chemical bonds within drug molecules. Besides, they typically resort to simple concatenation operations to achieve feature fusion.

To better extract the drug feature within the drug molecular, instead of using 1D-CNNs, some works tend to use GNNs to represent drug SMILES sequences as molecular topology graphs \cite{b12, b13, b14, b15}.
SGNetDTA \cite{b16} employs graph attention algorithms to extract drug features from molecular topology graphs and uses 1D-CNNs for protein feature extraction. ColdDTA \cite{b17} combines GNNs with dense layers, incorporating residual connections between each GNN to prevent information loss. 
However, these methods often neglect the global topology information. 

In this paper, we propose an enhanced DTA prediction framework ViDTA. We employ the Graph Transformer to extract features from drug molecules and introduce a virtual node to capture global features. Finally, the high-level features of proteins and drugs will be fed into a carefully designed attention-based linear feature fusion network for affinity prediction. 

The main contributions  are summarized  as follows:
\begin{itemize}
\item We introduce the virtual nodes to the Graph Transformer network for feature extraction, providing a broader receptive field and capturing richer global structural correlations between atoms in drug molecules.
\item We propose an attention-based linear feature fusion network that incorporates a gated skip connection mechanism, which can better capture interaction information between drug and protein features.
\item The experiments on multiple benchmarks with prevalent evaluation metrics demonstrate that ViDTA outperforms state-of-the-art baselines.
\end{itemize}

\section{Materials and Methodology}

\subsection{Datasets}

We evaluated our method over three public benchmark datasets (Davis \cite{b18}, Metz \cite{b19}, and KIBA \cite{b20}). The dataset details are provided in Table~\ref{tab1}. Smaller $K_d$ indicates a higher affinity between the drug and the target. To reduce the variance, $K_d$ values in the Davis dataset are typically transformed into logarithmic space. The transformation process is formulated as:
\begin{equation}
pK_d=-\lg \left( \frac{K_d}{10^9} \right) 
\label{eq1}
\end{equation}

\subsection{Overview}
\begin{figure*}[t]
\centering
\includegraphics[width=\textwidth]{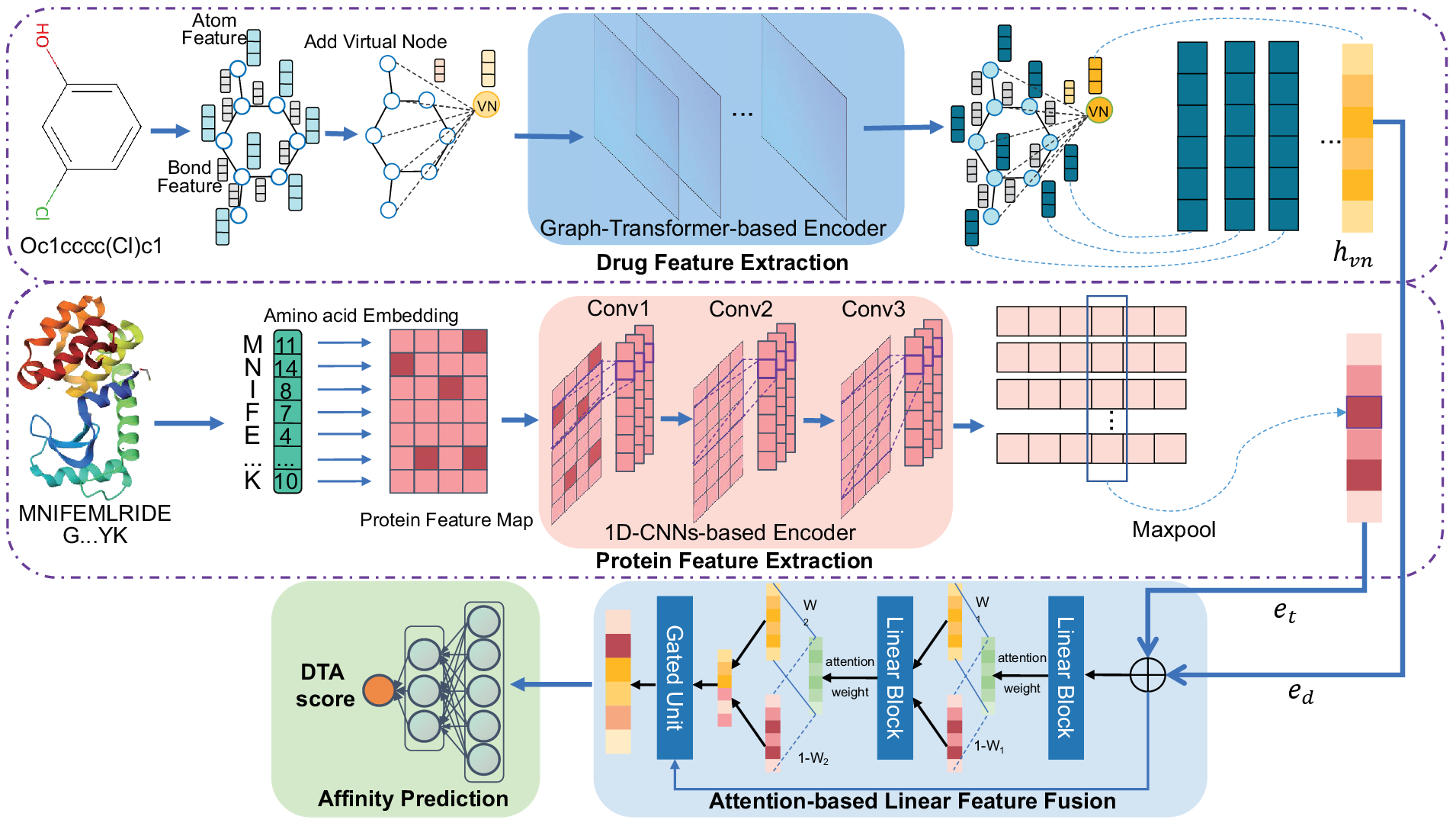}
\caption{Overview of the proposed ViDTA}
\label{fig1}
\end{figure*}

Our proposed ViDTA model consists of four modules: a drug feature extraction network, a protein feature extraction network, an attention-based Linear Feature Fusion network, and an affinity prediction network. The overview of  ViDTA is illustrated in Fig.~\ref{fig1}.

\begin{table}[t]
\centering
\caption{Statistical analysis of benchmark datasets}
\setlength{\tabcolsep}{14.0pt}
\scalebox{0.99}{
\begin{tabular}{c|c|c|c}
\toprule[0.15em]
\rowcolor{mygray} Dataset & Drugs& Proteins& Affinities \\
\midrule[0.1em]
Davis & 68& 442& 30056 \\
Metz & 240& 121& 13669\\
KIBA & 2111& 229& 118254 \\
\bottomrule[0.15em]
\end{tabular}}
\label{tab1}
\end{table}

The drug feature extraction network takes the drug molecule's SMILES sequence as input.
The SMILES sequence is transformed into a two-dimensional molecular graph of atoms and bonds.
Then we add a virtual node to the graph, which is fed to a Graph-Transformer-based encoder to extract the feature of the drug molecule. 

Concurrently, the protein feature extraction network processes the protein's FASTA sequence.
The protein sequence is encoded into a protein feature map using embedding vectors derived from various types of amino acids.
The protein feature map is fed into a 1D-CNNs-based encoder to extract the sequence feature of the protein.
The features extracted by both the drug and protein networks are then input into an attention-based linear feature fusion network to generate fused features. Finally, the fully connected layers are used to predict drug-target affinity based on the fused features.

\subsection{Drug Feature Extraction}
\subsubsection{Drug Molecular Graph Representation}
The drug feature extraction network takes the SMILES sequence as input. We first convert the SMILES sequence into a drug molecular graph $G=(V,E)$. In this graph, the set of nodes $V$ denotes the atom, while the set of edges $E$ represents the feature vectors for chemical bonds between atoms. 

We add a virtual node $vn$ to the graph $G$. The virtual node is connected to all atomic nodes, forming virtual edges. The initial feature vector of the virtual node and edges are initialized to zero and added to $V$ and $E$, respectively. Following  AttentionMGT-DTA \cite{b21}, each atom's initial feature vector is determined based on its properties such as symbol, formal charge, atom hybridization, and atom chirality, resulting in a 44-dimensional vector. The initial feature vector for each chemical bond is derived from other properties like bond type, aromatic or conjugated, resulting in a 10-dimensional vector.

\subsubsection{Graph-Transformer-bsed Drug Encoder}
\begin{figure}[t]
\centering
\hspace*{-0.05\columnwidth}
\includegraphics[width=\columnwidth]{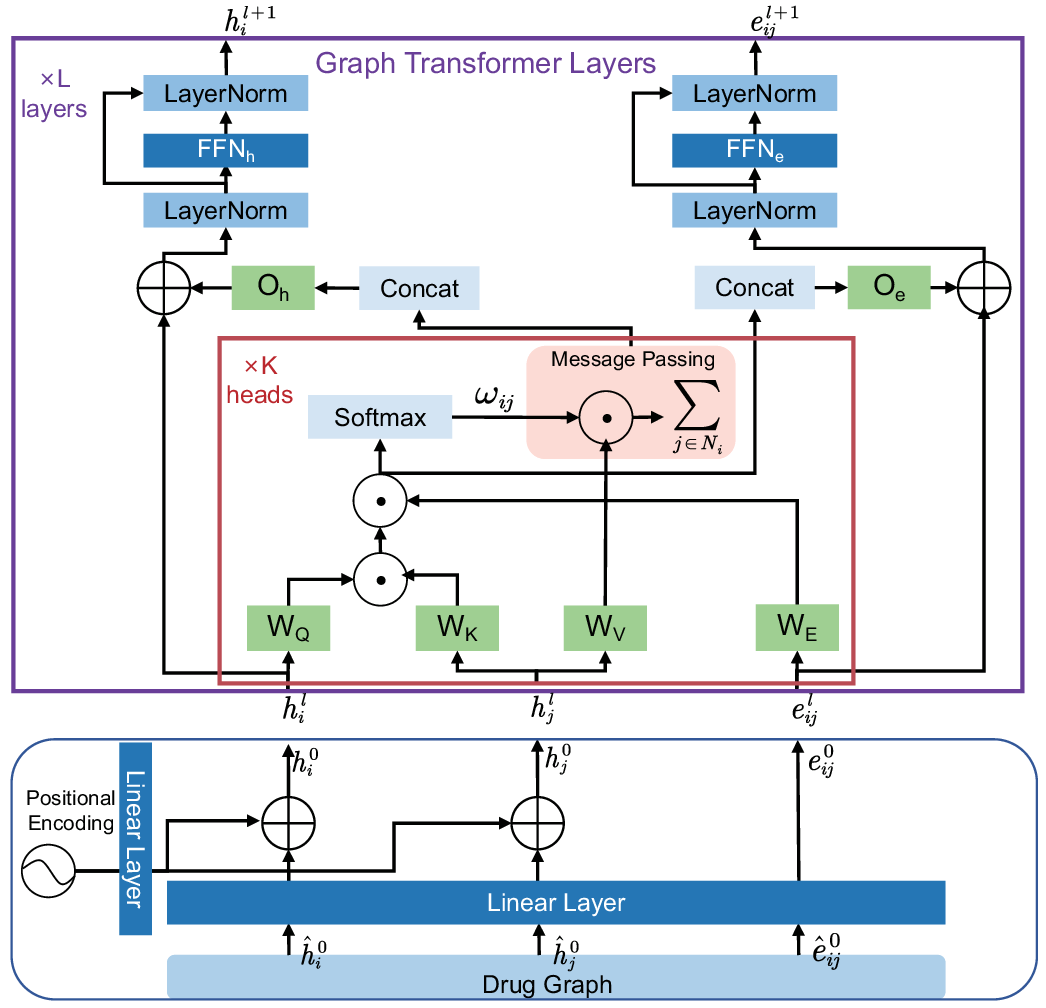}
\caption{Flowchart of drug encoder based on Graph Transformer}
\label{gt}
\end{figure}

Fig.~\ref{gt} illustrates the framework of the drug encoder based on Graph Transformer\cite{b22}. For the drug graph $G$, the initial atomic feature of the $i_{th}$ node, and the chemical bond feature of the edge between the $i_{th}$ node and the $j_{th}$ node, are first mapped through a linear layer to obtain the atomic feature $\hat{h}_{i}^{(0)}$ and the edge feature $e_{ij}^{(0)}$ of length $ d_k $, where $ d_k $ is the input dimension of drug encoder.

Subsequently, we calculated the symmetrically normalized Laplacian matrix $L$  using the identity matrix $I$, the degree matrix $D$, and the adjacency matrix $A$ of the drug molecular graph:
\begin{equation}
    L=I-D^{-\frac{1}{2}}AD^{-\frac{1}{2}}
    \label{eq2}
\end{equation}

The eigenvector of 
$L$ is mapped to $\lambda_i\in R^{d_k}$ through a linear layer.
Then $\lambda_i$ is added to the atomic feature $\hat{h}_{i}^{(0)}$ to obtain the input node feature $h_{i}^{(0)}$ for the Graph Transformer:
\begin{equation}
    h_{i}^{(0)}=\hat{h}_{i}^{(0)}+\lambda_i
    \label{eq3}
\end{equation}


The attention score $\hat{\omega}_{ij}^{(l)}$ between the $i_{th}$ node and $j_{th}$ node in the $l_{th}$ layer of the Graph Transformer is obtained as:
\begin{equation}
\hat{\omega}_{ij}^{(l)}=\frac{W_{Q}^{(l)}h_{i}^{(l)}\cdot W_{K}^{(l)}h_{j}^{(l)}}{\sqrt{d_h}}W_{E}^{(l)}e_{ij}^{(l)}
    \label{eq4}
\end{equation}
\begin{equation}
\omega _{ij}^{(l)}=\text{Softmax}\left( \hat{\omega}_{ij}^{(l)} \right) 
    \label{eq5}
\end{equation}
where $W_{Q}^{(l)}, W_{K}^{(l)} \in R^{d_h \times d_k}$ are the linear transformation matrices for the feature vectors of the $i_{th}$ and $j_{th}$ nodes in the $l_{th}$ layer, respectively, $W_{E}^{(l)} \in R^{d_h \times d_k}$ is the linear transformation matrix for the feature vector of the edge between the $i_{th}$ and $j_{th}$ nodes in the $l_{th}$ layer, and $d_h$ is the unified dimension of the hidden feature.

Then for the multi-head attention layer of the Graph Transformer, we utilize a message-passing neural network \cite{b23} to update the features of both nodes and edges.

For the node message passing, the feature of the $i_{th}$ node is updated by aggregating the current features of neighboring nodes (considered as the $j_{th}$ node), weighted by attention scores:
\begin{align}
\hat{h}_{i}^{(l+1)}&=\sum_{j\in N_i}{\omega _{ij}^{(l)}W_{V}^{(l)}h_{j}^{(l)}}
\label{eq6}
\end{align}
where $W_{V}^{(l)} \in R^{d_h \times d_k}$ is the linear transformation matrix of the feature vectors of neighboring nodes for the $i_{th}$ node in the $l_{th}$ layer.

For the edge message passing, $\hat{\omega}_{ij}^{(l)}$ is used as the new edge feature $\hat{e}_{ij}^{(l+1)}$:
\begin{align}
\hat{e}_{ij}^{(l+1)}&=\hat{\omega}_{ij}^{(l)}
    \label{eq7}
\end{align}

Subsequently, we concatenate all the outputs obtained from the $K$ attention heads, and then add them to the input features of the $l_{th}$ layer:
\begin{align}
\hat{\hat{h}}_{i}^{(l+1)}=O_{h}^{(l)}\underset{k=1}{\overset{K}{||}}\hat{h}_{i,k}^{(l+1)} +{h}_{i}^{(l)}\notag\\
  \hat{\hat{e}}_{ij}^{(l+1)}=O_{e}^{(l)}\underset{k=1}{\overset{K}{||}}\hat{e}_{ij,k}^{(l+1)}+{e}_{ij}^{(l)}
    \label{eq8}
\end{align}
where $||$ represents the concatenation operation, $O_{h}^{(l)} \in R^{K*d_h\times d_o}$ and $O_{e}^{(l)} \in R^{K*d_e\times d_o}$ denote the learnable weight matrices, $d_o$ is the output dimension of the drug encoder.

The final feature of nodes and edges in the $(l+1)_{th}$ layer are obtained using a feed-forward neural network (FNN) and residual modules:
\begin{align}
    {h}_{i}^{(l+1)}&=\text{Norm}\left(\hat{\hat{h}}_{i}^{(l+1)}+W_{h2}^{(l)}\text{ReLU}\left(W_{h1}^{(l)}\text{Norm}\left( \hat{\hat{h}}_{i}^{(l+1)}\right) \right) \right) 
    \label{eq9}
\end{align}
\begin{align}
{e}_{ij}^{(l+1)}&=\text{Norm}\left(\hat{\hat{e}}_{ij}^{(l+1)}+W_{e2}^{(l)}\text{ReLU}\left(W_{e1}^{(l)}\text{Norm}\left( \hat{\hat{e}}_{ij}^{(l+1)}\right) \right) \right) 
    \label{eq10}
\end{align}
where "Norm" refers to Layer Normalization, $W_{h}^{(l)}$ and $W_{e}^{(l)}$ are learnable parameters in $R^{d_o \times d_o}$.

The node feature of the last layer's virtual node $e_d={h}_{vn}^{(L)}$ is regarded as the final representation of the molecular graph.

\subsection{Protein Feature Extraction}

\subsubsection{Protein Sequence Representation}
Target proteins are biological macromolecules composed of multiple amino acids, each represented by a unique letter. According to the work of Tsubaki et al.~\cite{b24}, each of the 25 amino acids is assigned an integer value (e.g., {``A": 1, ``C": 2, ``B": 3, \textit{etc.}}). To standardize input dimensions, we set the maximum length of protein sequences to 1000. Sequences longer than this are truncated, and sequences shorter are padded with zeros. 

\subsubsection{1D-CNNs-based Protein Encoder}
Each integer representing an amino acid is mapped to a unique $ d_p $ dimensional vector through an embedding layer. This results in an input protein feature map $M_p\in R^{l_p\times d_p} $, where $ l_p $ is the maximum length of amino acid sequences, and $ d_p $ is the size of the protein embedding. 

We then use three concatenated 1D-CNN blocks as the protein encoder to obtain the protein feature map of each layer. To maintain the consistency of input and output sequence lengths and to increase the receptive field, we used progressively larger convolutional kernel sizes (2, 3, 5), with padding sequentially set to 5, 7, and 11. The stride and dilation were consistently set to 1. The final representation of the protein sequence $e_t$ is obtained by performing a max pooling operation. It can be performed as follows:
\begin{equation}
e_t=\text{MaxPool}\left(\text{Conv1d}({M_p}) \right) 
\label{eq11}    
\end{equation}
\subsection{Attention-based Linear Feature Fusion}
\begin{figure}[t]
    \centering
    \hspace*{-0.05\columnwidth}
    \includegraphics[width=\columnwidth]{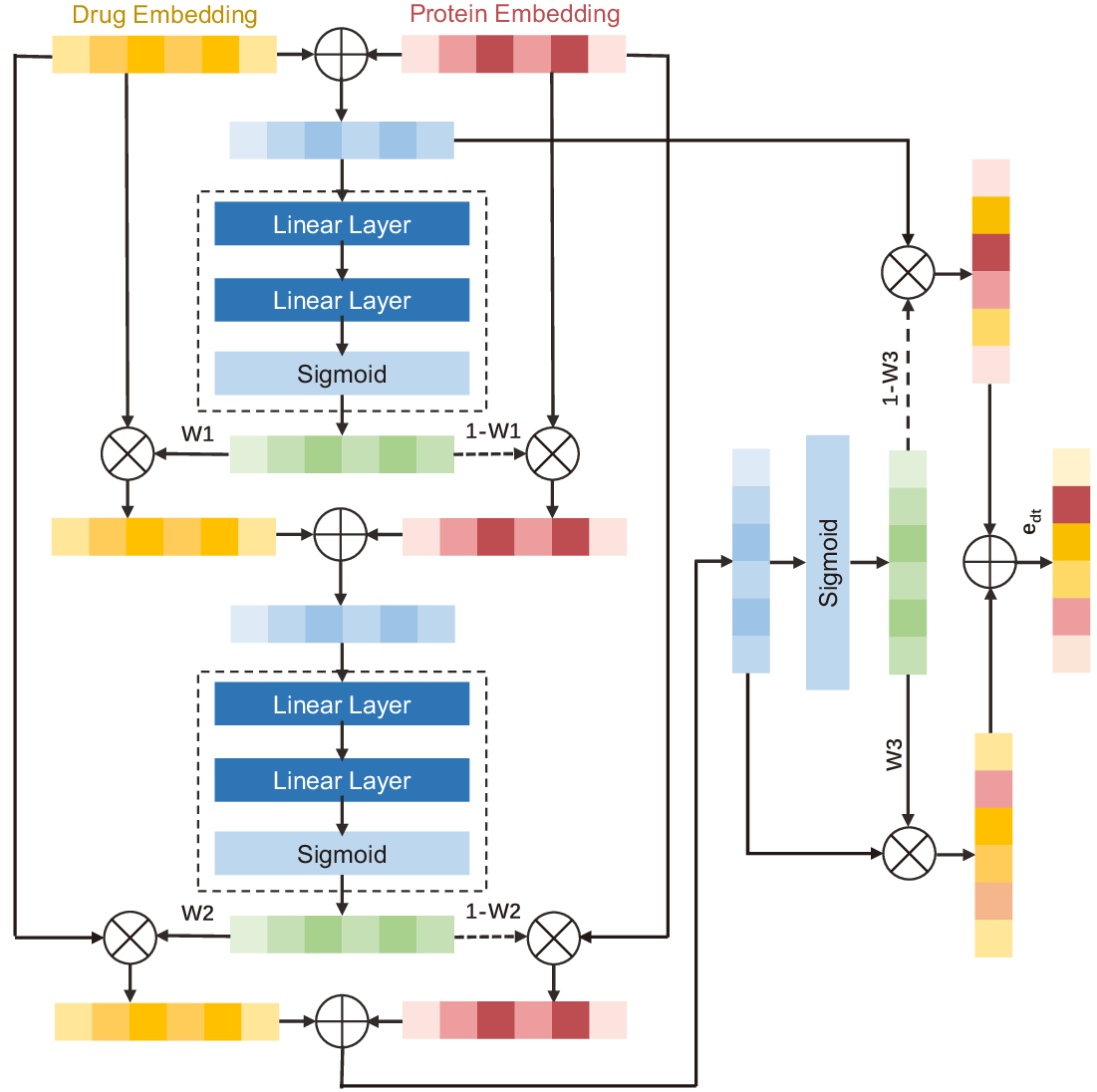}
    \caption{Flowchart of the attention-based linear feature fusion network}
    \label{LFFA}
\end{figure}
After the above processing, the drug feature $e_d$ and protein feature $e_t$ are fed into the attention-based linear feature fusion network. The overall workflow of the feature fusion module is illustrated in Fig.~\ref{LFFA}. Firstly, $e_d$ and $e_t$ are added into a block composed of two linear layers and a sigmoid activation function to obtain the attention weight score $W_1$:
\begin{equation}
    W_1=\text{Sigmoid}\left(\text{Linear}\left(\text{Linear} \left(e_d+e_t\right)\right)\right)
    \label{eq12}
\end{equation}

$W_1$ and $(1-W_1)$ are then respectively multiplied by the drug and target protein feature to get the fused feature $e_{dt}^{1}$: 
 \begin{equation}
     e_{dt}^{1}=W_1*e_d+ \left(1-W_1 \right)*e_t
     \label{eq13}
 \end{equation}
 
This process is repeated by inputting $e_{dt}^{1}$ into the second linear block to get the second weight score $W_2$, and the second fused feature $e_{dt}^{2}$.
Finally, using a gated skip connection mechanism \cite{b25} which integrates features from different hidden layers, the initial features $e_d$ and $e_d$, and the fused feature $e_{dt}^{2}$ are weighted to get the final fused representation of the drug and protein target. In this way, we can preserve primary features, ensuring that critical information is not overlooked during training:
\begin{equation}
    W_3=\text{Sigmoid}\left(e_{dt}^{2}\right)
    \label{eq14}
\end{equation}
\begin{equation}
    e_{dt}=W_3*e_{dt}^{2}+\left(1-W_3\right)*\left(e_d+e_t\right)
    \label{eq15}
\end{equation}

\subsection{Affinity Prediction Network}
The affinity prediction network is comprised of four fully connected layers, which take the fused feature $e_{dt}$ as input. Besides, Batch-Normalization and ReLU activation functions are applied between adjacent linear layers.

The loss function used for model training is MSE Loss:

\begin{equation}
    \text{MSE}=\frac{1}{N} \sum\left(\hat{y}_i-y_i\right)^2
    \label{eq16}
\end{equation}
where $N$ is the number of samples, $\hat{y}_i$ represents the predicted affinity value, and $y_i$ is the true affinity value.

\begin{table}[!t]
\centering
\caption{Hyperparameter settings of ViDTA}
\setlength{\tabcolsep}{8.0pt}
\scalebox{0.99}{
\begin{tabular}{c | c }
\toprule[0.15em]
\rowcolor{mygray} Hyper-parameters & Setting \\
\midrule[0.15em]
Learning rate & \{3e-4,1e-4\} \\
Batch size& 128 \\
Epoch& 500\\
Dimension of the drug encoder input layer&128\\
Dimension of the drug encoder hidden layers& 16\\
Dimension of the drug encoder output layer& 128\\
Dimension of the protein encoder input layer& 128\\
Dimension of the protein encoder hidden layer&256\\
Dimension of the protein encoder output layer& 128\\

Layers number of the Graph Transformer & 10\\
Heads number of the Graph Transformer &8 \\
Dropout of Graph Transformer layers & 0.2\\
\bottomrule[0.1em]\end{tabular}}
\label{tab2}
\end{table}

\begin{table}[!t]
\centering
\caption{comparison results on the Davis benchmark dataset}
\setlength{\tabcolsep}{8.0pt}
\scalebox{0.99}{
\begin{tabular}{c | c  |c | c  |c}
\toprule[0.15em]
\rowcolor{mygray} Method & CI $\uparrow$ & $r_{m}^{2}$ $\uparrow$ & PCC$\uparrow$ & MSE $\downarrow$\\
\midrule[0.15em]
GraphDTA	&0.888	&0.699	&0.8475	&0.232\\
GOaidDTA	&0.891	&0.654	&0.850	&0.229\\
rzMLP-DTA	&0.896	&0.709	&-	&0.205\\
AttentionDTA	&0.8947	&0.7404	&0.8721	&0.1912\\
ColdDTA	&0.8938&	0.7606	&0.8861	&0.1695\\
TF-DTA&	0.8856	&0.6703	&-&0.2312\\
TEFDTA	&0.8925	&0.7403	&	0.8617&0.2100\\
DGDTA	&0.899&	0.702	&-&0.225\\
TransVAE-DTA&	0.8696	&0.5713	&-&0.3329\\
AttentionMGT-DTA	&0.891	&0.699&	-&0.193\\
\midrule
ViDTA (ours)	&\textbf{0.9052}&	\textbf{0.7654}&	\textbf{0.8882}&	\textbf{0.1680}\\
\bottomrule[0.1em]\end{tabular}}
\label{tab3}
\end{table}

\begin{table}[!t]
\centering
\caption{comparison results on the Metz benchmark dataset}
\setlength{\tabcolsep}{8.0pt}
\scalebox{0.99}{
\begin{tabular}{c | c  |c | c  |c}
\toprule[0.15em]
\rowcolor{mygray} Method & CI $\uparrow$ & $r_{m}^{2}$ $\uparrow$ & PCC$\uparrow$ & MSE $\downarrow$\\
\midrule[0.15em]
GraphDTA	&0.8621	&0.7079	&0.8548&	0.1714\\
ArkDTA	&0.8430	&	-	& -&0.1703\\
AttentionDTA	&0.8755&	0.7048	&0.8565	&0.1612\\
ColdDTA	&0.8738	&0.7116	&0.8622&	0.1553\\
TEFDTA	&0.8445	&0.6171&	0.8376	&0.1873\\
\midrule
ViDTA (ours)	&\textbf{0.8848}&	\textbf{0.7526}	&\textbf{0.8729}	&\textbf{0.1434}\\
\bottomrule[0.1em]\end{tabular}}
\label{tab4}
\end{table}

\begin{table}[t]
\centering
\caption{comparison results on the KIBA benchmark dataset}
\setlength{\tabcolsep}{8.0pt}
\scalebox{0.99}{
\begin{tabular}{c | c  |c | c  |c}
\toprule[0.15em]
\rowcolor{mygray} Method & CI $\uparrow$ & $r_{m}^{2}$ $\uparrow$ & PCC$\uparrow$ & MSE $\downarrow$\\
\midrule[0.15em]
AttentionDTA	& 0.8799& 	0.7350	& 0.8739	& 0.1668\\
GOaidDTA	& 0.876	& 0.706	& 0.868	&0.179\\
rzMLP-DTA	&0.890	&0.748	&-	&0.142\\
ColdDTA	& 0.8689	& 0.7054& 	0.8671	& 0.1762\\
TEFDTA	& 0.8675	& 0.7065	& 0.8546	& 0.1864\\
TF-DTA	& 0.8768& 	0.7344& 	-& 	0.1771\\
TransVAE-DTA	& 0.8221	& 0.6329	& -& 0.2536\\
AttentionMGT-DTA& 	0.893& 	0.786& 	-& 0.140\\
\midrule
ViDTA (ours)	& \textbf{0.8981}& 	\textbf{0.7932}	& \textbf{0.8989}& 	\textbf{0.1347}\\
\bottomrule[0.1em]\end{tabular}}
\label{tab5}
\end{table}

\begin{table}[!t]
\centering
\caption{Ablation results on different drug representation}
\setlength{\tabcolsep}{8.0pt}
\scalebox{0.99}{
\begin{tabular}{c | c  |c | c  |c}
\toprule[0.15em]
\rowcolor{mygray} Drug representation & CI $\uparrow$ & $r_{m}^{2}$ $\uparrow$ & PCC$\uparrow$ & MSE $\downarrow$\\
\midrule[0.15em]
GCN	&0.8585	&0.6705	&0.8388	&0.1793\\
GAT&	0.8606	&0.6975&	0.8524	&0.1653\\
GIN	&0.8658	&0.7204&	0.8653	&0.1519\\
Graph Transformer	&0.8667 & 0.7205 & 0.8646   & 0.1523\\
Transformer	&0.8285	&0.5984	&0.8063&	0.2081\\
\midrule
\makecell{Graph Transformer with\\ virtual node (ours)} &\textbf{0.8848}	&\textbf{0.7526}	&\textbf{0.8729}	&\textbf{0.1434}\\
\bottomrule[0.1em]\end{tabular}}
\label{tab6}
\end{table}

\begin{table}[!t]
\centering
\caption{Ablation results on different feature fusion}
\setlength{\tabcolsep}{8.0pt}
\scalebox{0.99}{
\begin{tabular}{c | c  |c | c  |c}
\toprule[0.15em]
\rowcolor{gray!30} Feature Fusion & CI $\uparrow$ & $r_{m}^{2}$ $\uparrow$ & PCC$\uparrow$ & MSE $\downarrow$\\
\midrule[0.15em]
Addition	&0.8699	&0.7192	&0.8669	&0.1503\\
Concatenation &	0.8759	&0.7479&	0.8726	&0.1440\\
\midrule
\makecell{Attention-based  \\ linear feature fusion \\ network (ours)}  &\textbf{0.8848}	&\textbf{0.7526}	&\textbf{0.8729}	&\textbf{0.1434}\\
\bottomrule[0.1em]\end{tabular}}
\label{tab7}
\end{table}

\section{Experiment and Result}
\subsection{Experiment setting}
To ensure the reliability of our experimental results, we employed five-fold cross-validation in this study. To enhance the training efficiency, we initialize the learning rate at 0.0003 and decay it to 0.0001 after completing 100 epochs. The batch size for all three benchmark datasets was uniformly set to 128, with training continuing for 1000 epochs. If the model's loss did not decrease after 200 epochs, training was halted to prevent overfitting. Additional details of hyperparameters can be found in the Table~\ref{tab2}.

\subsection{Evaluation Metrics}
DTA prediction is a regression task, therefore, we use Concordance Index (CI)~\cite{b26}, Modified Correlation Coefficient ($r_{m}^{2}$)~\cite{b27}, Pearson Correlation Coefficient (PCC)~\cite{b28}, and Mean Squared Error (MSE) as evaluation metrics.

\subsection{Experimental Results}

We compared our approach with several methods, including GraphDTA \cite{b12}, GOaidDTA \cite{b29}, rzMLP-DTA \cite{b30}, AttentionDTA \cite{b10}, coldDTA \cite{b17}, TF-DTA \cite{b31}, TEFDTA \cite{b11}, DGDTA \cite{b32}, TransVAE-DTA \cite{b33}, and AttentionMGT-DTA \cite{b21}, ArkDTA \cite{b34}.

The experimental results illustrated in Table~\ref{tab3}, Table~\ref{tab4} and Table~\ref{tab5} demonstrate that the ViDTA model performed outstanding in these three benchmark datasets. Among the four metrics of CI, $r_{m}^{2}$, PCC, and MSE, ViDTA stands out as the top-performing model.

Notably, our results have several significant highlights: In the Davis dataset, our model achieved a CI index of 0.9052, while other baselines didn't score higher than 0.9. In the Metz dataset, the MSE of our scheme decreased by 7.7\% from 0.1553 to 0.1434. In the KIBA dataset, PCC increased by 2.9\% from 0.8739 to 0.8989.

\subsection{Ablation Experiments}
We compared the impact of different drug feature extraction architectures on DTA prediction in the Metz dataset, including GCN, GAT, GIN, Graph Transformer based on molecular graphs, and Transformer based on SMILES sequences. We also evaluated the prediction performance of virtual nodes versus readout as molecular graph representations. The experimental results are presented in Table \ref{tab6}, which indicate that incorporating virtual nodes further enhanced the performance across all metrics.

Additionally, we assessed the influence of different feature fusion methods. By replacing the proposed attention-based linear feature fusion network with simple addition or concatenation, the experimental results demonstrate that our approach consistently outperformed the other two methods across all metrics, as shown in Table \ref{tab7}.

\section{Conclusion}

In this study, we propose ViDTA, an enhanced drug-target affinity prediction scheme. By introducing the virtual nodes to the Graph Transformer network, our method shows significant potential in simultaneously extracting local and global features from the drug molecular graph. Furthermore, the attention-based linear feature fusion network that incorporates a gated skip connection mechanism effectively integrates the features from both drug and protein targets. Our approach has demonstrated remarkable performance on multiple widely used benchmark datasets. The experiments with prevalent evaluation metrics demonstrate that ViDTA outperforms state-of-the-art baselines.








\vspace{12pt}

\end{document}